\def\year{2022}\relax
\documentclass[letterpaper]{article} 
\usepackage{aaai22}  
\usepackage{times}  
\usepackage{helvet} 
\usepackage{courier}  
\usepackage[hyphens]{url}  
\usepackage{graphicx} 
\usepackage{natbib}  
\usepackage{caption} 
\usepackage[utf8]{inputenc}
\usepackage{amsmath}
\usepackage{amsthm}
\usepackage{adjustbox}
\usepackage{color}
\usepackage{amsmath,amssymb,amsfonts}
\usepackage{algorithmic}
\usepackage{graphicx}
\usepackage{textcomp}
\usepackage{xcolor}
\def\BibTeX{{\rm B\kern-.05em{\sc i\kern-.025em b}\kern-.08em
    T\kern-.1667em\lower.7ex\hbox{E}\kern-.125emX}}
    
\def\mod     {{\textit{CRSDnet}}}

\mathchardef\mhyphen="2D
\newcommand*{\rom}[1]{\expandafter\@slowromancap\romannumeral #1@}

\title{Opinion Spam Detection: A New Approach Using Machine Learning and Network-Based Algorithms}
\author{
    Kiril Danilchenko\thanks{Supported by ISF grant number 1388/16.},
    Michael Segal and
    Dan Vilenchik
}
\affiliations{
    School of Electrical and Computer Engineering\\

    Ben-Gurion University of the Negev\\
    Beer-Sheva, Israel 8410501\\
    vilenchi@bgu.ac.il
    
}

\begin{document}
\maketitle

\begin{abstract}
E-commerce is the fastest-growing segment of the economy. Online reviews play a crucial role in helping consumers evaluate
and compare products and services. As a result, fake reviews (opinion spam) are becoming more prevalent and negatively impacting customers and service providers. There are many reasons why it is hard to identify opinion spammers automatically, including the absence of reliable labeled data. This limitation precludes an off-the-shelf application of a machine learning pipeline. We propose a new method for classifying reviewers as spammers or benign, combining machine learning with a message-passing algorithm that capitalizes on the users' graph structure to compensate for the possible scarcity of labeled data. We devise a new way of sampling the labels for the training step (active learning), replacing  the typical uniform sampling. 
Experiments on three large real-world datasets from Yelp.com show that our method  outperforms  state-of-the-art active learning approaches and also machine learning methods that use a much larger set of labeled data for training.
\end{abstract}

\section*{Introduction}\label{section:introduction}
In the era of e-commerce, consumers typically buy products or services based on reviews.  Therefore reviews are increasingly valuable for sellers and service providers due to the benefits of positive reviews or the damage from negative ones. In light of this, fake reviews are flourishing and pose a real threat to the proper conduct of e-commerce platforms. A group of reviewers (we will call them opinion spammers) post fake reviews to promote their products or demote their competitors' products.

Fake reviews are often written by experienced professionals who are paid to write high-quality, believable reviews.
Detecting opinion fraud is a non-trivial and challenging problem that was extensively studied in the literature using various approaches. Some approaches are based solely on the review content \cite{jindal2008opinion,li-etal-2013-topicspam,ott-etal-2011-finding}, reviewer behavior \cite{lim2010detecting,mukherjee2013yelp}
and the tripartite relationships between reviewers, reviews, and products \cite{10.1145/2187836.2187863,rayana2015collective,rayana2016collective,wang2011review,10.1007/s10115-017-1068-7}. While each paper presented a method that is useful to some extent for detecting  certain kinds of spamming
activities, there is no one-size-fits-all solution. This is because spammers keep changing their strategies, many times in an adaptive manner to the spam detection policies. Therefore, there is a need to study and incorporate
as many approaches as possible.

Another challenge is the fact that most datasets are imbalanced, as the three datasets that we used for evaluation, where only about 20\% of the users are spammers (Table \ref{tab:modVSgnn}).

Despite the vast commercial impact of opinion spam detection, most machine-learning-based solutions for this problem do not achieve very high performance due to insufficient labeled data to properly train an ML model. In addition, standard ML models often treat each sample separately, disregarding the underlying graph structure of the spammer group. Indeed, this graph is often latent and should be derived from the data in an ad-hoc manner.
One can use deep graph learning methodology to automatically embed the users, and thus infer the underlying graph structure, but such an approach requires a large training  set, which is often not available or costly to obtain.

The setting where only part of the data is labeled is often called  ``semi-supervised learning". When very few examples are labeled, this setting is also termed  ``few-shot learning". When in addition one is given the possibility to choose which set of users will be labeled (given a budget, one can decide for which users to invest the budget in order to acquire their label), this is called the active-learning setting. In this paper we study the few-shot active learning  setting for the opinion spam  detection problem.

The few-shot active learning setting is very reasonable for the opinion-spam detection problem because labeled data never comes labeled for free, and operators of e-commerce websites choose de-facto which users they want to check manually and label, and typically only a small fraction of the users are labelled. 


  
  \begin{figure}[h]
     \includegraphics[width=0.55\textwidth]{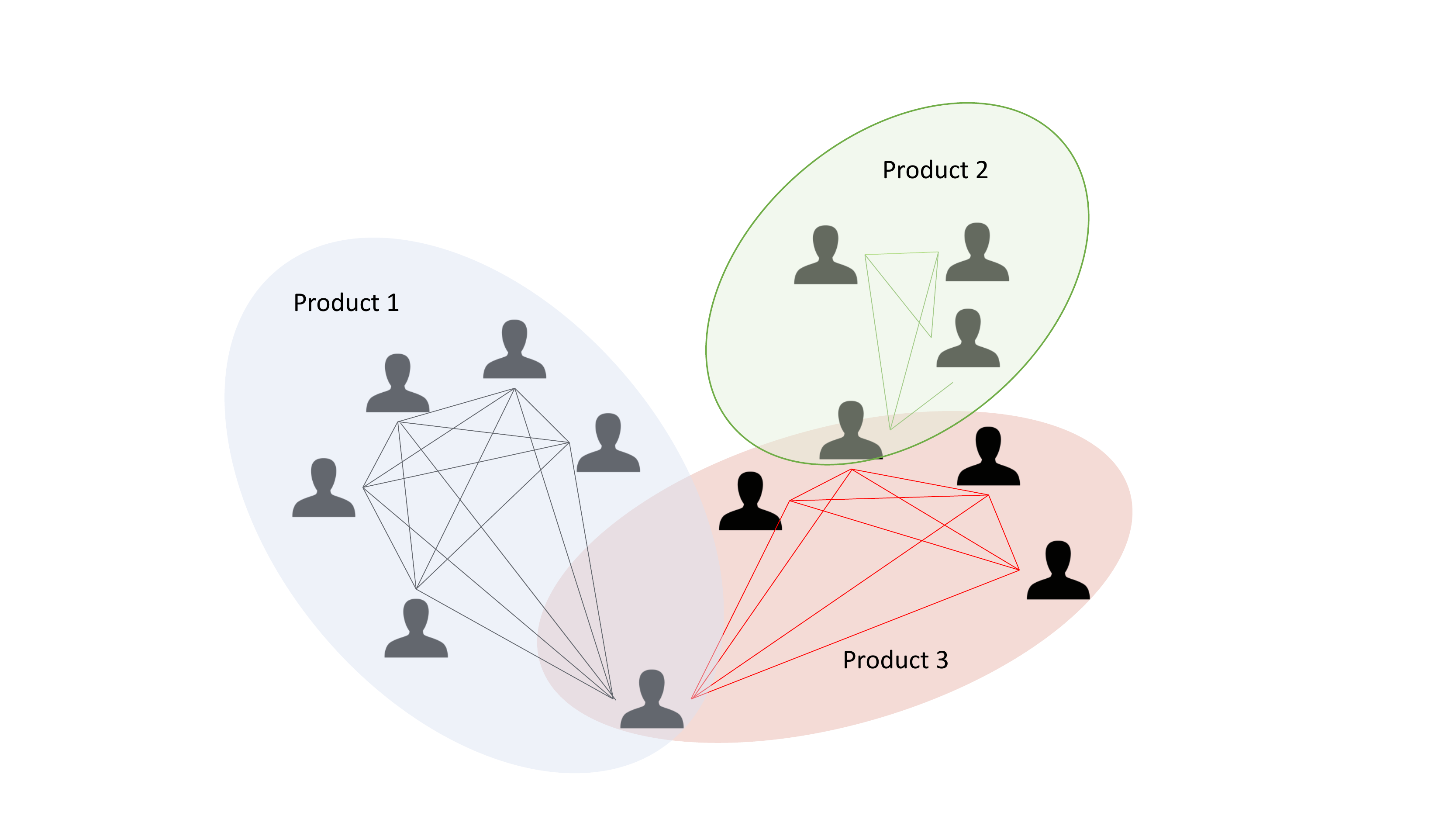}
     \caption{Illustration of the user-user graph, inducing an interconnected set of cliques, one for each product.}
     \label{fig:cliques}
 \end{figure}
\subsection*{Our Contribution}
We propose a classification algorithm, {\em Clique Reviewer Spammer Detection Network}, \mod , for detecting fake reviews in the few-shot active learning setting. \mod~  harnesses the power of both machine learning algorithms and classical graphical models algorithms such as Belief Propagation. We show that this combination yields a better performance than each approach separately. 

We evaluate our algorithm on a golden-standard dataset for the spammer detection task, the Yelp Challenge Data \cite{YelpData}. The performance of our algorithm is better than all previous work in almost every metric. We also outperform other methods that use the graph structure, such as graph embedding, and neural graph networks algorithms~\cite{liu2020alleviating,wang2019fdgars,9435380}. These algorithms use much more labeled data for training (30\% and over, compared to at most 2.5\% that we use). 

We show that using both machine learning and the graph structure (via label propagation algorithms)  improves over stand-alone machine learning by at least 10\% in AUC measure. 

We attribute the success of our algorithm to the following key innovations in our approach:
\begin{itemize}
    \item  We derive from the raw data a user-user graph, where two users share an edge if they wrote a review for the same item. Figure \ref{fig:cliques} illustrates this graph, which is a clique graph by definition. In previous work a tripartite  user-review-product graph was used. 
    \item The user-user graph may be much denser than the tripartite user-review-product graph. To overcome computational issues that such density entails, we design a careful edge sparsification procedure to speed up the algorithm without compromising performance much. The sparsification is guided by the rule that each node will end up having just enough edges connecting it to nodes both from his own class (spammer or not) and from the opposite class.
    \item We run a label-propagation algorithm (concretely, Belief Propagation), but some parameters of that algorithm (the node and edge potentials) are determined using a machine learning model. This is the first time that such a combination of approaches is undertaken, and its usefulness demonstrated.
    \item We propose a new way of choosing the set of users whose labels will be obtained (active learning). Instead of randomly choosing a set of users up to the allowed budget, we choose random users from the largest clique of the user-user graph. The intuition behind this rule comes from the work of Wang et. al. \cite{wang2020collueagle} where it was shown that collusive spamming (or, co-spamming) is a useful lens to identify spammers.
\end{itemize}

\section*{Related Work}\label{section:rw}
Opinion spam detection has different nuances such as fake review detection \cite{jindal2008opinion,ott-etal-2011-finding,10.1145/2187980.2188164,10.1145/1281192.1281280}, fake reviewer detection \cite{lim2010detecting,wang2011review,10.1145/2505515.2505700} and spammer group detection \cite{10.1145/2187836.2187863, 10.1007/978-3-319-23528-8_17, wang2020collueagle,10.1007/s10115-017-1068-7,10.1007/s10489-018-1142-1}. Two survey papers, \cite{Crawford2015SurveyOR,VivianiSurvey},  provide a broad perspective on the field.

Our method is part of the graph-based models, which take into account the relationships among reviewers, comments, and
products. The key algorithm in this approach is Belief Propagation (BP)~\cite{pearl2014probabilistic} which is applied to a carefully designed graph and the Markov Random Field (MRF) associated with it.
The first to use this approach were Akoglu et al. \cite{akoglu2013opinion} who suggested FraudEagle, a BP-based algorithm that runs on the bipartite reviewer-product graph, where the edge potentials are based on the sentiment in the review. In later work, Rayana et. al. \cite{rayana2015collective} introduced SPEagle, where node and edge potentials  are derived from a richer set of meta-data features, improving significantly over the performance of FraduEagle. Wang et al. \cite{wang2011review} consider the tripartite user-review-product network and define scores for trustiness of users, honesty of reviews, and reliability of products. They use an ad-hoc iterative procedure to compute the scores, rather than BP. In \cite{7865975} an algorithm called NetSpam was introduced which utilizes spam features for modeling review datasets as heterogeneous information networks.

A graph-based approach was suggested in \cite{fei2013exploiting} but this time the graph contains edges for reviews that were written within a certain time difference from each other (a ``burst"). The authors use a different dataset to evaluate their method, reviews from Amazon.com.

\begin{figure*}[h]
  \centering
  \includegraphics[width=0.9\linewidth]{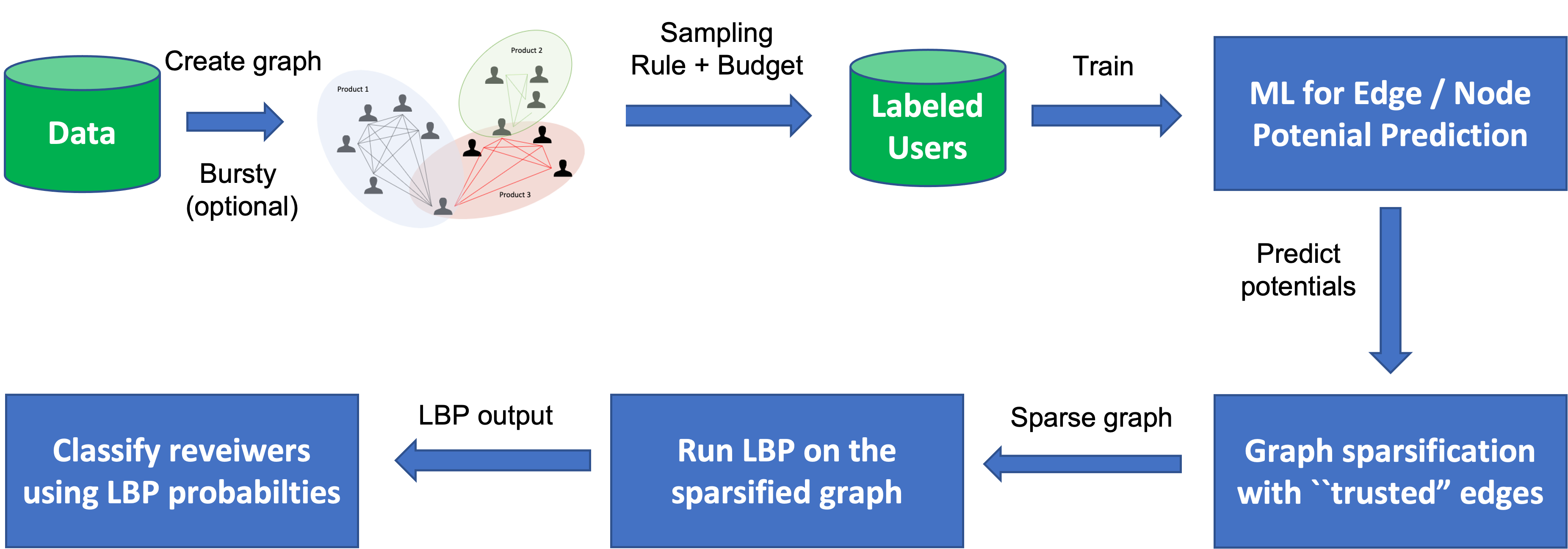}
  \caption{The flow chart of our pipeline, from raw data, through sampling users for labeling, using them to train an ML algorithm for predicting edge and node potentials, sparsification of the graph using trusted edges, running LBP, and completing the classification task.}
  \label{fig:flowChart}
\end{figure*}

The authors of \cite{wang2020collueagle} present ColluEagle, a graph-based algorithm to detect both review spam campaigns and {\em collusive} review spammers. To measure the collusiveness of pairs of reviewers, they identify reviewers that review the same product in a similar way.

A different approach to the problem of spammer detection is via Graph Neural Networks (GNNs) and Graph Convolutional Networks (GCNs). These are deep learning architectures for graph-structured data. The core idea is to learn node representations through local neighborhoods~\cite{kipf2016semi}. 
In  \cite{liu2020alleviating}, the authors design a new GNN framework, GraphConsis, to
tackle the fraud detection task. The authors evaluated the method on four data sets, where one of them is used by us too. GraphConsis is benchmarked on different training set sizes, from 40\% to 80\% of the data. GraphConsis with $80\%$ training set size achieves AUC of $0.742$ on the Yelp Chicago dataset, and our method achieves  $0.754$ with only $2.5\%$. Note that GraphConsis uses all the metadata that we use as well (the graph structure, and the reviews). 

A  GCN-based algorithm was designed in \cite{10.1145/3308560.3316586}, and tested on reviews from Tencent Inc. The algorithm outperformed four baseline algorithms (Logistic Regression, Random Forest, DeepWalk~\cite{perozzi2014deepwalk}, LINE~\cite{tang2015line}).

Our work departs from previous work in several ways. Compared to the works where label-propagation algorithms were used, we use machine learning to predict the edge and node potentials rather than hand-crafted threshold  functions. Second, we consider the user-user graph and not a bi/tri-partite user-review-product graph. To overcome the computational challenge incurred by the density of the user-user graph, we apply a new rule for edge sparsification, based on the ML prediction. Finally, in the active learning setting, we introduce a new sampling rule. All these modifications have led to an improvement over FraudEagle \cite{akoglu2013opinion} and SPEagle \cite{rayana2015collective}.

\noindent\textbf{Active Learning: }
The active learning approach aims to achieve high accuracy by using few queries, and therefore the ``most informative" points are natural candidates for label acquisition. Various heuristics were proposed to determine the ``most informative" nodes, e.g., uncertainty sampling \cite{lewis1994heterogeneous,culotta2005reducing,settles2008analysis} and variance reduction \cite{flaherty2006robust,schein2007active}. In our setting, we chose a rule that is native to the problem itself -- sampling from the largest clique, following the take-home message from the work of \cite{wang2020collueagle} about collusive spamming.

Many works on active learning choose the train set using  adaptive rules, point by point. This however is infeasible in our case as re-running the entire pipeline for every new example is computationally prohibitive for datasets as large as ours. Therefore we choose all users for labelling in bulk.

\section*{Methodology} \label{sec:PropMetho}
In this section we describe our pipeline, end to end. The flow chart is depicted in Figure \ref{fig:flowChart}.

We formulate the spam detection problem as a classification task on the user network. The dataset 
consists of  $n$ reviewers who write reviews on $m$ products from the set $P$. 
The vertex set of the  graph $G=(V,E)$ is the set of users (reviewers); user $i$ and $j$ share an undirected edge if there exists some product $p \in P$ such that user $i$ and $j$ wrote a review for $p$.  The resulting graph consists of interconnected cliques, each corresponds to a different product. Figure \ref{fig:cliques} illustrates such a network. Each node $i \in V$ has in addition a vector of features $F_i$ associated with it, and a binary class variable  $v_i \in \{1,-1\}$, for spammer (1) or benign (-1).
 
The classification task is given the graph $G$ (with the nodes' features), and possibly a set of labeled users $\{i_{1},\dots i_{k}\}$ (the ``few shots" training set),  predict the value of $v_i$ for the remaining nodes (the test set).

Ideally, to solve the classification task, we would  find an assignment $s:\{-1,1\}^n \to V$ that maximizes
\begin{align}\label{eq:OrigProb}
  Pr\left[ v_1 = s_1,\ldots, v_n = s_n | v_{i_1}=s_{i_1}, \ldots, v_{i_{k}}=s_{i_{k}},G\right].  
\end{align}

This Maximum Likelihood Estimation (MLE) task is in general NP-hard. However, in practice, a useful solution $s$ (perhaps not the maximizer) may be obtained by using a Markov Random Field modelling for the probability space.
\subsection*{Markov Random Field}

Markov Random Field (MRF) is often used to model a set of random variables having a Markov property described by an undirected dependency graph. A pairwise-MRF (pMRF) is an MRF satisfying the pairwise Markov property: a random variable depends only
on its neighbors and is independent of all other variables.

A pMRF model involves two types of potentials, node potentials, and edge potentials.
Our node potentials, $\phi_i(v_i)$, stand for the probability that reviewer $i$ belongs to either class (spam/benign):
\begin{align}\label{eq:nodePotential}
   \phi_i(v_i)=\begin{cases}
a_i & , v_i=1 \ \ (i \text{ is a spammer})\\
1-a_i & ,v_i=-1 \ \ (i \text{ is a benign).}
\end{cases}\\
\notag 
\end{align}
The edge potential $\psi_{ij}(v_{i},v_{j})$ signifies the affinity of reviewer $i$ and $j$, namely, the probability $p_{ij}$ that both belong to the same class. Formally,
\begin{align}\label{eq:edgePotential}
   \psi_{ij}(v_i,v_j)=\begin{cases}
p_{ij} & ,v_i=v_j\\
1-p_{ij} & ,v_i \neq v_j.
\end{cases}\\
\notag 
\end{align}
The parameters $a_i$ and $p_{ij}$ satisfy $a_i,p_{ij} \in [0,1]$ for all $i,j$. To determine the values of these parameters we use machine learning applied to features that are extracted from the metadata of the reviewers dataset. 

The pMRF model is used to approximate the expression for $Pr[s|G]$ in Eq.~\eqref{eq:OrigProb}:
\begin{align}\label{eq:objectiveFunction}
   \Pr\left(s\right)=\frac{1}{Z}\prod_{v_i \in V}\phi_i(v_{i})\prod_{\left( i,j\right) \in E}\psi_{ij}(v_{i},v_{j}),
\end{align}
 
 where $Z$ is a normalization factor, the sum over the energies of all possible $2^{|V|}$ assignments $s$.

Finding the assignment $s$ that maximizes the probability in Eq.~\eqref{eq:objectiveFunction} is still an intractable problem;  LBP (loopy belief propagation) is the go-to heuristic for approximating the intractable maximization problem.

The LBP algorithm \cite{pearl2014probabilistic} is based on an iterative message passing along the edges of the graph. Messages are initialized according to some user-defined rule. At iteration $t$, a message $m^{(t)}_{ij}$ is sent from node $i$ to each neighboring node $j$. The message represents the belief of $i$ about the label of $j$. If $G$ is a tree, then BP is guaranteed to converge; if $G$ contains cycles then convergence is not guaranteed (hence the name loopy), but in practice, a cap on the number of iterations is set. We use the standard LBP messages, omitted for brevity, and can be found in \cite{pearl2014probabilistic}.

Each iteration of LBP takes $O(|V|+|E|)$ time, hence the number of edges, which may be quadratic in $|V|$, plays a key role in the computational complexity. The more iterations one can perform for the same time budget, the better the performance. In the next section, we describe how to address the computational aspect using graph sparsification.

\subsection*{Running LBP on a Sparse Graph}\label{sec:method:sparsification1}
Recall that our graph is defined over users, and not as a tri-partite user-product-review graph. This may result in a rather dense graph, which poses a computational impediment even on LBP, when the number of nodes is large.  
For example, the graph created from the Yelp Chicago dataset is very dense (average degree 1193). Therefore our first step is to sparsify the graph by choosing a linear number of ``useful" edges (linear in the number of nodes). 

To gain intuition into a useful way of sparsification, we conducted the following experiment using the Chicago Yelp data \cite{rayana2015collective}. The initial graph contains $38063$ nodes, and,  $2.4 \cdot 10^7$, edges. The sparsification procedure is parameterized with  two numbers $k_{1}$ and $k_{2}$. For every node $i$ we choose $k_{1}$ neighbors from $i$'s class (spammer or benign) and $k_{2}$ neighbors from the opposite class, and color these $k_1+k_2$ edges red.  We then remove all edges that were not colored red. The resulting graph has an average degree of at most $k_1+k_2$ (multiple edges are merged).  We set all node potentials to  $\phi(v_{i})\gets \{0.5,0.5\}$; in other words, we don't provide any prior knowledge about the class of the node $v_i$. We set all edge potentials $\psi_{ij}$ as follows: $p_{ij}=\epsilon$ if $v_i \ne v_j$ and $p_{ij}=1-\epsilon$ if $v_i = v_j$ (that is, according  to the true agreement relationship between users $i$ and $j$).  We fix  $\epsilon=0.001$.

We run LBP on the resulting graph, and label each node $v_i$ as a spammer if the probability that LBP assigns it is larger than a pre-defined threshold $\tau$.

   \begin{figure}[t!]
  \centering
  \includegraphics[width=0.9\linewidth]{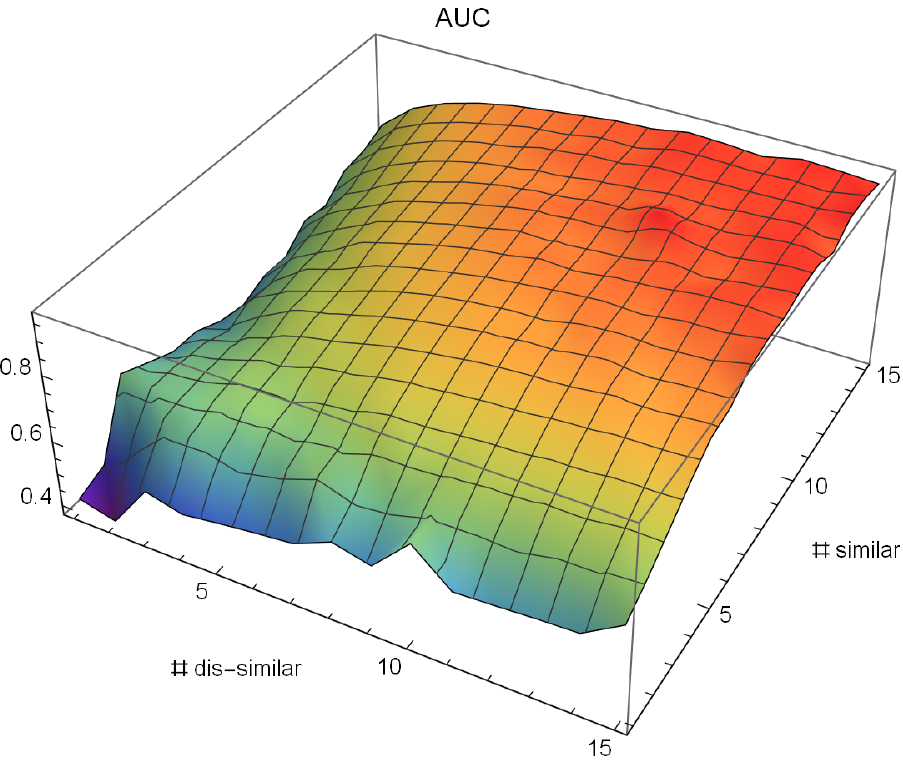}
  \caption{The AUC of the LBP algorithm for the Yelp Chicago dataset. LBP is run on a subgraph in which each node has $k_1$ neighbors from the similar class and $k_2$ neighbors from the other (dis-similar) class. Edge potentials are set according to the ground truth.}
  \label{fig:edgeMotivation}
\end{figure}

Figure  \ref{fig:edgeMotivation} depicts the AUC of the LBP classification when varying $k_1$ and $k_2$ between 0 and 15. We see that the AUC approaches 1 when each node has ``enough" neighbors from each class.

 \subsection*{Sparsification and Edge Potentials}\label{section:method:edge_potentials}
In practice, we have the true labels of a small set of nodes, and we use it to learn and predict the edge potentials between the remaining unlabelled nodes. We shall use this prediction for a sparsification procedure similar to what was just described. Our sparsification proceeds as follows:

(1) The first step is to train a machine learning algorithm on the set of users whose labels we know with the objective of predicting $p_{ij}$, the probability that a pair of reviewers $i,j$ belongs to the same class (either both spammers or both benign). The exact choice of ML algorithm, along with the parameters is explained in experimental setting section.  (2) Compute $p_{ij}$ using the ML model for all the remaining edges of the graph (edges that do not connect two users from the training set). (3) Choose all edges for which $p_{ij} \in [0.95,1]$ or $p_{ij} \in [0,0.05]$ and set the potential in Eq.~\eqref{eq:edgePotential} accordingly. We call these edges the ``trusted" edges. LBP will be run on the graph containing the trusted edges.

The input to the machine learning algorithm in steps (1) and (2) is a set of features that is extracted from the metadata of both the users and the reviews. Typical metadata includes the text of the review, the rating that the reviewer gave the product, the total number of  reviews that the reviewer wrote, etc. The Yelp set of features is described in Tables \ref{tab:User_Features} and \ref{tab:Review_Features}.

Additional sparsification of the graph can be obtained by  removing edges that connect users whose reviews of the same product were written in faraway times. Namely, two users $i$ and $j$ share an edge only if they wrote a review for the same product, and these reviews were written within a period of $T$ days (following previous work, we fixed $T=7$). Such a graph with time-dependent edges is called a { \em bursty} graph and was introduced in \cite{fei2013exploiting}. We tested our pipeline with and without the bursty variant. The bursty sparsification, if applied, is done before the trusted edges are selected.

\subsection*{Node Potential}\label{sec:method:node_potential}
In the experiment just described, all node potentials were set to $\{0.5,0.5\}$, and  only  edge potentials played a role. However, there may be a gain in setting the node potentials according to the metadata features rather than ignoring it.

Similar to the way we set the edge potential, we use machine learning to predict $a_i$ in Eq.~\eqref{eq:nodePotential}. The machine learning algorithm is trained on the set of users chosen for labeling (active learning setting) with the objective of predicting spam or benign. The value of $a_i$ is predicted for all the remaining users using the trained model (which gives the ``probability" of being a spammer or benign, alongside the discrete label).

\subsection*{Active learning: Sampling Users}\label{sec:method:sampling}
The final component in our methodology is the way we choose the set of users for training. In this work we explore two sampling rules with the same budget of $k$ users:
\begin{enumerate}
 \item \textit{Random Sampling:} Pick $k$ reviewers uniformly at random from $V$. \label{stra:RS}
\item \textit{Sampling from largest clique:} \label{stra:LargestClique}
In this strategy, we sample $k$ users that belong to the largest clique. The largest clique corresponds to the product on which the largest number of reviews were written. If the budget is not consumed, we sample the remainder from the second-largest clique, and so on.

\end{enumerate}

 \begin{table}[t!]
\centering
    \begin{tabular}{ |p{1.0cm}|p{1.7cm}|p{2.0cm}|p{1.59 cm}| }
  \hline
     Dataset &   \#Reviews \newline(fake \%) &  \#Users \newline(spammer \%) & \#Products\\
      \hline
        \hline
     Y'Chi       &67,395 \newline(13.23 \%)     & 38,063 \newline(20.33\%) &201 \\
      \hline
     Y'NYC       & 359,052 (10.27\%)   &160,225 (17.79\%)  &923  \\
      \hline
     Y'Zip       &   608,598 (13.22\%) &260,277 (23.91\%)  &5,044   \\   
          \hline
\end{tabular}
 \caption{Summary statistics of the three Yelp datasets \cite{mukherjee2013yelp,rayana2015collective}.}
\label{tab:modVSgnn}
\end{table}

\section*{Data Description}\label{section:Data}
To evaluate our methodology  we use three datasets that contain reviews from  Yelp.com, summary statistics of which are presented in Table \ref{tab:statDS}. The datasets contain reviews of restaurants and hotels and were collected by \cite{mukherjee2013yelp,rayana2015collective}. YelpChi covers the Chicago area,  YelpNYC covers NYC and YelpZip is the largest, and it includes ratings and reviews for restaurants in a continuous region including NJ, VT, CT,  PA, and NY. They differ in size (YelpChi is the smallest and YelpZip is the largest), as well as in the percentage of spammers out of the total number of users. Yelp has a filtering algorithm that identifies fake/suspicious reviews. The three datasets contain these labels. We partition the users into spammers: authors of at least one filtered review and benign: authors with no filtered reviews.

Alongside the text of the reviews, the dataset contains additional metadata such as ratings, timestamps. From the text and the additional data, various features are extracted, which were used in previous work that studied these datasets \cite{rayana2015collective,mukherjee2013yelp,lim2010detecting}. Tables \ref{tab:User_Features} and \ref{tab:Review_Features}  include brief descriptions of these features. Most of them are self-explanatory, and hence we omit detailed explanations for brevity. Note that we used exactly the same set of features as \cite{rayana2015collective} to allow a fair comparison.

The features are used to compute both the node and the edge potentials as explained in the Methodology section.

\begin{table}[h]
\centering
    \begin{tabular}{ |p{0.7cm}|p{6.5cm}|}
  \hline
 \multicolumn{2}{|c|}{User Features} \\
      \hline
     MNR         & Max. number of reviews written in a day \cite{mukherjee2013spotting,mukherjee2013yelp}    
          \\\hline
     PR          &  Ratio of positive reviews (4-5 star) \cite{mukherjee2013yelp}  \\\hline
     NR           &Ratio of negative reviews (1-2 star) \cite{mukherjee2013yelp}      \\\hline
          avgRD           &Avg. rating deviation of  user’s reviews \cite{mukherjee2013yelp,lim2010detecting,fei2013exploiting}  
         \\ \hline
     WRD           &Weighted rating deviation \cite{lim2010detecting}\\\hline
     BST           & Burstiness \cite{mukherjee2013yelp,fei2013exploiting} (spammers are often short-term members of the site).   
\\\hline
     RL           &Avg. review length in number of words \cite{mukherjee2013yelp}      \\\hline
     ACS           &
Avg. content similarity—pairwise cosine similarity among user’s (product’s) reviews, where a review is represented as a bag-of-bigrams \cite{lim2010detecting,fei2013exploiting}    \\\hline
     MCS           &Max. content similarity—maximum cosine similarity among all review pairs \cite{mukherjee2013spotting}    
    \\ \hline

\end{tabular}
 \caption{User Features }
    \label{tab:User_Features}
\end{table}
 \begin{table}[h]
\centering
    \begin{tabular}{ |p{0.7cm}|p{6.5cm}|}
  \hline
 \multicolumn{2}{|c|}{Review Features} \\

      \hline
     Rank         & Rank order among all the reviews of product \cite{jindal2008opinion}     \\\hline
     RD          &  Absolute rating deviation from product’s average rating \cite{li2011learning}  \\\hline
     EXT           &Extremity of rating \cite{mukherjee2013spotting}   \\\hline
     DEV           & Thresholded rating deviation of review  \cite{mukherjee2013spotting}
\\\hline
     ETF           & Early time frame \cite{mukherjee2013spotting} (spammers often review early to increase impact)
          \\\hline
    ISR           & If review is user’s sole review, then $x_{ISR} =1$, otherwise 0 \cite{rayana2015collective}   \\\hline
     PCW           &Percentage of ALL-capitals words  \cite{li2011learning,jindal2008opinion}  \\\hline
     PC           &Percentage of capital letters \cite{li2011learning} \\\hline
     L           &Review length in words \cite{li2011learning}    \\\hline
    PP1         &Ratio of 1st person pronouns (‘I’, ‘my‘, etc.) \cite{li2011learning}\\\hline
    RES         &Ratio of exclamation sentences containing ‘!’ \cite{li2011learning}\\
     \hline

\end{tabular}
 \caption{Review Features}
    \label{tab:Review_Features}
\end{table}
 \section*{Evaluation}\label{section:Simulation}
In this section, we describe the results of the experiments we ran on the three Yelp datasets. We report our results and results obtained by previous work on the same datasets.
\subsection*{Evaluation Metrics}
We evaluated the performance of \mod~using four popular metrics, which were used by previous work as well. The Average Precision (AP), which is the area under the precision-recall curve, the ROC AUC, and the precision@k. To compute precision@k we rank the reviewers according to the probability that LBP assigned each one to be a spammer. We compute the fraction of real spammers among the top $k$ places. We compute precision@k  for $k = 100, 200,\dots, 1000$.

Finally, we use the Discounted Cumulative Gain (DCG@k) which
provides a weighted score that favors correct spammer predictions at the top indices. Formally, $DCG@k=\sum_{i=1}^k=\frac{2^{l_i}-1}{\log_2(i+1)}$ where $l_i=1$ if the user at the $i^{th}$ place is a correctly identified spammer, and 0 otherwise. For compatibility with other works, we actually report the normalized $DCG$, which is obtained by dividing $DCG@k$ by the ideal $DCG$ which is the $DCG@k$ where all $l_i=1$ (all top $k$ are indeed spammers in the ideal ranking, for the $k$ that we choose).

  \begin{table}[ht!]
\centering
    \begin{tabular}{ |c|c|c|c|c|}
  \hline
    Setting  &  Nodes & Edges  & Sampling & Bursty   \\ \hline\hline
    1 &         ML   & None   & Random & No \\ \hline
    2 &         ML   & Threshold  & Random & No  \\ \hline
    3 &         Threshold & ML & Random & No \\ \hline
    4 &         ML & ML &  Random & No \\ \hline
    5 &         ML & ML &  Clique & No \\ \hline
    6 &         ML & ML &  Random & Yes \\ \hline
    7 &         ML & ML &  Clique & Yes \\ \hline
    
\end{tabular}
 \caption{Various configurations of running  \mod. The first two columns say how potentials were computed: using ML or the threshold method of \cite{rayana2015collective}. The sampling rule corresponds to the two options mentioned at the end of the Methodology section. Bursty refers to time-dependent sparsification. Setting \#1 consists of only using ML to predict nodes class, without LBP.}
\label{tab:Methods}
\end{table}

\begin{table*}[ht!]
\centering

    \begin{tabular}{|p{1.3cm} |p{0.75cm}|p{0.75cm}|p{0.75cm}|p{0.75cm}||p{0.75cm}|p{0.75cm}|p{0.75cm}|p{0.75cm}||p{0.75cm}|p{0.75cm}|p{0.75cm}|p{0.75cm}|  }
  \hline
 Method &  \multicolumn{4}{c||}{Y'Chi } & \multicolumn{4}{c||}{Y'NYC  } & \multicolumn{4}{c|}{Y'Zip }\\
  \hline
              &$0.25\%$     & $0.5\%$  &  $1\%$    & $2.5\%$  & $0.25\%$     & $0.5\%$ &  $1\%$    & $2.5\%$  & $0.25\%$     & $0.5\%$ &  $1\%$    & $2.5\%$    \\
          \hline
SpEagle   &                &               & 0.691          &                                                                &        &        & 0.657       &     &  &        & 0.671       &               \\
\hline
SpEagle$^+$  &               &               & 0.708          &                                                                &   &               & 0.683       &  & & & 
0.691&       \\
\hline
NetSPAM   &                &   &                &                                                        &        &        &0.650       &0.650     &        &        &        &        \\
\hline\hline
Set. 1 & 0.519                &0.602                &0.620                 & 0.632                                                          &       0.561 &        0.578&       0.558 &   0.585&        0.566&       0.593 &        0.672&0.693        \\

\hline
Set. 2 & 0.669          & 0.701       & 0.711        & 0.729                                                        & 0.664 & 0.663 & 0.687 &    0.692 & 0.685 & 0.700 & 0.784 & 0.794 \\
\hline
Set. 3 & 0.688          & 0.699        & 0.702          & 0.702  & 0.659 & 0.677  & 0.691 &    0.692 & 0.562 & 0.708 & 0.779 & 0.831 \\
\hline
Set. 4 & 0.689        & 0.712         & 0.723         & 0.731                                                          & 0.673 & 0.681 &       0.685 &    0.665 & 0.684 & 0.707 & 0.706 & 0.828 \\
\hline
Set. 5 & {\bf 0.718}          & {\bf 0.724}         & {\bf 0.735}       & {\bf 0.754}  & {\bf 0.669} & { \bf 0.688} & {\bf 0.720} &    {\bf 0.766} & {\bf 0.703} & {\bf 0.729} & {\bf 0.790} & {\bf 0.848} \\
\hline
Set. 6 & 0.673  & 0.730 & 0.719  & 0.741  & 0.668 & 0.671 & 0.682 &     0.696& 0.628 & 0.707 & 0.783 & 0.835 \\
\hline

Set. 7 & 0.672          &                0.694 & 0.662          & 0.668                                                          & 0.666 & 0.666 &       0.645 &     0.645& 0.631 & 0.726 & 0.763 & 0.824\\
 \hline
\end{tabular}
\caption{AUC performance of compared methods on all three Yelp datasets. Best results are in bold. Empty cells stand for results that were not reported in the paper or were not computed by us. }
    \label{tab:AUC}

\end{table*}
 \begin{table*}[ht!]
 \centering
    \begin{tabular}{|p{1.3cm} |p{0.75cm}|p{0.75cm}|p{0.75cm}|p{0.75cm}||p{0.75cm}|p{0.75cm}|p{0.75cm}|p{0.75cm}||p{0.75cm}|p{0.75cm}|p{0.75cm}|p{0.75cm}|  }
  \hline
 Method &  \multicolumn{4}{c||}{Y'Chi } & \multicolumn{4}{c||}{Y'NYC  } & \multicolumn{4}{c|}{Y'Zip }\\
  \hline
              &$0.25\%$     & $0.5\%$  &  $1\%$    & $2.5\%$  & $0.25\%$     & $0.5\%$ &  $1\%$    & $2.5\%$  & $0.25\%$     & $0.5\%$ &  $1\%$    & $2.5\%$    \\
          \hline
         
\hline
SpEagle$^+$ &        &       & 0.396     &         &        &         &  0.348       &         &      &        & 0.424      &     \\
\hline
NetSPAM  &       &       &      &         &         &         & ~0.300        &~0.28         &       &        &       &      \\
\hline\hline
Set. 1   & 0.874        & 0.802      & 0.896     & 0.852     & 0.902       & 0.913        &0.917        &0.916        & 0.885     &0.896      & 0.901      & 0.886     \\
\hline
Set. 2   & 0.901  &        0.879& 0.882 & 0.906    & 0.915 & 0.912 & 0.921 & 0.912&        0.782 &       0.805 &      0.859  & 0.883      \\
\hline
Set. 3   & 0.901  &       0.793 & 0.897 & 0.883    & 0.912 & 0.919 & 0.924 & 0.924 & {0.794}   & 0.845   & 0.923  & 0.941 \\
\hline
Set. 4   & 0.890  &     0.825   & 0.896 &         0.901 & 0.917 & {\bf0.922} & {\bf0.968}& 0.927 & {0.859}   & {0.870}   & 0.869  & 0.875 \\
\hline
Set. 5& {\bf0.909}&        {\bf0.906}& {\bf0.900} & {\bf0.914}&  0.875        &  0.727& 0.925& {\bf0.926}& {\bf0.903}& {\bf0.906}& {\bf0.935}&0.942 \\
\hline
Set. 6   & 0.907  &       0.707 & 0.886 & 0.913    & {\bf0.914} & 0.920 &         0.926 & 0.921 & {0.864}   & {0.891}   & 0.927  & {\bf0.946} \\
\hline
Set. 7   & 0.885  &       0.873 & 0.886 & 0.883 & 0.914 & 0.920 &        0.948  & 0.914 & {0.899}   & {0.833}   & 0.847  & 0.870\\
 \hline
\end{tabular}
 \caption{AP performance of compared methods on all three datasets. The best results are in bold. Empty cells stand for results that were not reported in the paper or were not computed by us.}
    \label{tab:AP}
\end{table*}

\begin{center}

  \begin{table*}[]
  \centering
\begin{tabular}{|l|l|l|l|l||l|l|l|l||l|l|l|l|l|l|}
\hline
     & \multicolumn{4}{l||}{Y’Zip}                                                        & \multicolumn{4}{l||}{Y'NYC}                                                        & \multicolumn{4}{l|}{Y’Chi}                                                        \\
     \hline
$k$    &  \rotatebox{90}{FraudEagle} & \rotatebox{90}{Wang} &  \rotatebox{90}{SpEagle$^+$} & \rotatebox{90}{\mod} &
 \rotatebox{90}{FraudEagle} & \rotatebox{90}{Wang}    & \rotatebox{90}{SpEagle$^+$} & \rotatebox{90}{\mod} 
& \rotatebox{90}{FraudEagle} & \rotatebox{90}{Wang} & \rotatebox{90}{SpEagle$^+$} & \rotatebox{90}{\mod}  \\
\hline
100    & 0.30       & 0.21      & 0.93   & {\bf1}      & 0.21       & 0.15     & 0.96   & {\bf0.98}   & 0.55       & 0.18     & 0.90     & {\bf0.99} \\\hline
200    & 0.30       & 0.19      & 0.81   & {\bf1 }     & 0.19       & 0.19     & {\bf0.96}   & 0.91   & 0.52        & 0.18    & 0.91     & {\bf0.99} \\\hline
300    & 0.38       & 0.21      & 0.69   & {\bf0.93}   & 0.17       & 0.18     & {\bf0.95}   & 0.86   & 0.48       & 0.20     & 0.91     & {\bf0.99} \\\hline
400    & 0.33       & 0.26      & 0.61   & {\bf0.80}   & 0.21       & 0.17     & {\bf0.95}   & 0.86   & 0.49       & 0.20     & 0.92     & {\bf0.99} \\\hline
500    & 0.29       & 0.27      & 0.57   & {\bf0.75}   & 0.22       & 0.17     & {\bf0.95}   & 0.88   & 0.48       & 0.20     & 0.92     & {\bf0.93} \\\hline
600    & 0.28       & 0.27      & 0.56   & {\bf0.74}   & 0.27       & 0.17     & {\bf0.96}   & 0.89   & 0.47       & 0.21     & 0.92     & {\bf0.90} \\\hline
700    & 0.27       & 0.29      & 0.54   & {\bf0.76}   & 0.37       & 0.16     & {\bf0.95}   & 0.90   & 0.47       & 0.21     & {\bf0.92}   & 0.91 \\\hline
800    & 0.26       & 0.30      & 0.51   & {\bf0.73}   & 0.45       & 0.16     & 0.90   & {\bf0.91}   & 0.49       & 0.22     & 0.91     & {\bf0.92} \\\hline
900    & 0.26       & 0.30      & 0.50   & {\bf0.69}   & 0.5        & 0.15     & 0.85   & {\bf0.92}   & 0.48       & 0.22     & 0.91     & {\bf0.92} \\\hline
1000   & 0.28       & 0.32      & 0.49   & {\bf0.67}   & 0.45       & 0.16     & 0.82   & {\bf0.92}   & 0.47        & 0.22    & 0.90    & {\bf0.93}\\
 \hline
\end{tabular}
\caption{Precision@k of compared methods when using 1\% of the users for training. The best results are in bold. \mod~runs in setting 5.}
    \label{tab:KP}
\end{table*}
\end{center}

  \begin{table}[ht!]
\centering
    \begin{tabular}{ |p{1.5cm} |p{3.5cm}|p{1.5cm}|}
  \hline
     Dataset &  Method &    AUC\\
      \hline
        \hline
     Y’Zip     &DFraud (80\%)~\cite{9435380} & 0.733  \\
      \hline
       Y’Zip     &RF (30\%)  & 0.740  \\
      \hline
        Y’Zip       &\mod~(2.5\%) & {\bf 0.847}    \\
      \hline
      \hline
      Y'Chi        &GraphConsis~\cite{liu2020alleviating} (80\%) &0.742    \\   
          \hline
           Y'Chi        &RF (30\%) &0.735   \\   
          \hline
          Y'Chi        &\mod~(2.5\%) & { \bf 0.754}   \\   
          \hline
\end{tabular}
 \caption{comparison with NN-based algorithms and a RF baseline. The percentage of data used for training appears in parenthesis.}
\label{tab:statDS}
\end{table}

\subsection*{The Experimental Setting}\label{sec:experiment}
There are four choices that effect the performance of \mod: (a) the way the node potentials are computed, using ML  or using a threshold function as in \cite{rayana2015collective}; (b) the way the edge potentials are computed, again using ML or using a threshold function~\cite{rayana2015collective}; (c) the active learning sampling rule which specifies how to choose the users for which the label is revealed; (d) with the time-dependent bursty sparsification or without. 

In Table \ref{tab:Methods} we summarize the seven different configurations with which we tested \mod. Each configuration was tested with a labeled set of users that is of size $0.25\%,0.5\%,1\%, 2.5\%$ of the entire set of users. In total we have $7 \times 4= 28$ experiments, each was ran 10 times with fresh randomness.

The machine learning algorithm that we used to compute the edge and node potentials was random forest, written in the Wolfram Language. The code is available on Github \footnote{\url{https://github.com/users/KirilDan/projects/1}}. We chose this implementation as Wolfram has a good support for graph structures, on which LBP can be easily run. All the parameters of the random forest are the default ones besides the following: 950 trees , a maximum tree depth of 16, and a maximum of $0.65$-fraction of the features are considered per split.
The features that we used are the ones in Tables \ref{tab:User_Features} and \ref{tab:Review_Features}.

To measure the extent to which each new component in our pipeline is responsible for the improvement over previous results, we  ran our pipeline also with the way that edge and node potentials were computed in \cite{rayana2015collective}. For completeness, we describe this method briefly. This method is completely unsupervised. A set of features $F_1,\ldots,F_r$ is computed for every user and review. Let $f_{u,i}$ be the value of feature $i$ for user $u$. For every feature $F_i$, the probability $p_{u,i}=Pr[F_i < f_{u,i}]$ is estimated from the data. Finally, a ``spam score" $S_u$ is computed for every user $u$ via
\begin{align}
    S_u =1-\sqrt{\frac{\sum_{i=1} ^r p_{u,i}}{r}}\label{eq:girls}
\end{align}
The potential of reviewer $v_{i}$ is set to $\phi(v_{i}) \leftarrow {\{1-S_u,S_u\}} $. A similar procedure is carried out to determine edge potentials.

\subsection*{Results}
Tables \ref{tab:AUC},\ref{tab:AP} and \ref{tab:KP} present the results of running \mod~ according to the aforementioned experimental setting, reporting the different evaluation metrics.

We compared the performance of \mod\ to other algorithms that were evaluated on the same dataset: SpEagle$^+$ \cite{rayana2015collective}, FraudEagle \cite{akoglu2013opinion}, NetSpam \cite{7865975}, Wang et al. \cite{wang2011review} and  ColluEgale \cite{wang2020collueagle}. The results that we report are taken from the relevant papers and are not reproductions that we carried out.

Different papers report different metrics and for different budgets. Hence some cells in the tables are left empty. Some algorithms are completely missing from a table/plot if the paper did not report that metric at all.

Table  \ref{tab:AUC} provides a comparison using the AUC measure. As evident from the table, our method is superior to all previous work. The most interesting comparison is with SpEagle+ \cite{rayana2015collective}. That work reports only results for a budget of 1\% of the users. For all three datasets, we already obtain a better result than SpEagle+ when using only 0.25\% of the users (6\% better for the Chicago dataset, 10\% better for the NYC dataset and 17\% for the ZIP).

Table \ref{tab:AP} reports the AP measure. Here the difference is even more dramatic. Our results are between 2 to 3 times better than SpEagle+ on all three datasets. 

Table \ref{tab:KP} reports the precision@k measure when using 1\% of the users and our best configuration, \#5. For the Chicago and ZIP dataset, our algorithm has the upper hand; for NYC, SpEagle+ outperforms \mod~for $k=200$ up to $k=700$, but the difference is very small. When our algorithm outperforms the other competitors, it is by a very large margin in most cases.

Tables \ref{tab:AUC} and \ref{tab:AP} suggest that the best way to run \mod~is according to configuration \#5 in Table \ref{tab:Methods}. Namely, both edge and node potentials are set using ML, and the budget is spent on users from the largest clique. Comparing settings 2,3 vs 4 we see that using ML for both nodes and edges (setting 4) is preferable to using ML only on one of them (settings 2,3). Settings 6 and 7 show that adding the time-dependent aspect, bursty edges, only harms the performance. Configuration \#1, using only ML applied to the user's features, gave the worse performance in the AUC measure, with a big gap.

\begin{figure}[htbp!]
  \centering
  \includegraphics[width=0.9\linewidth]{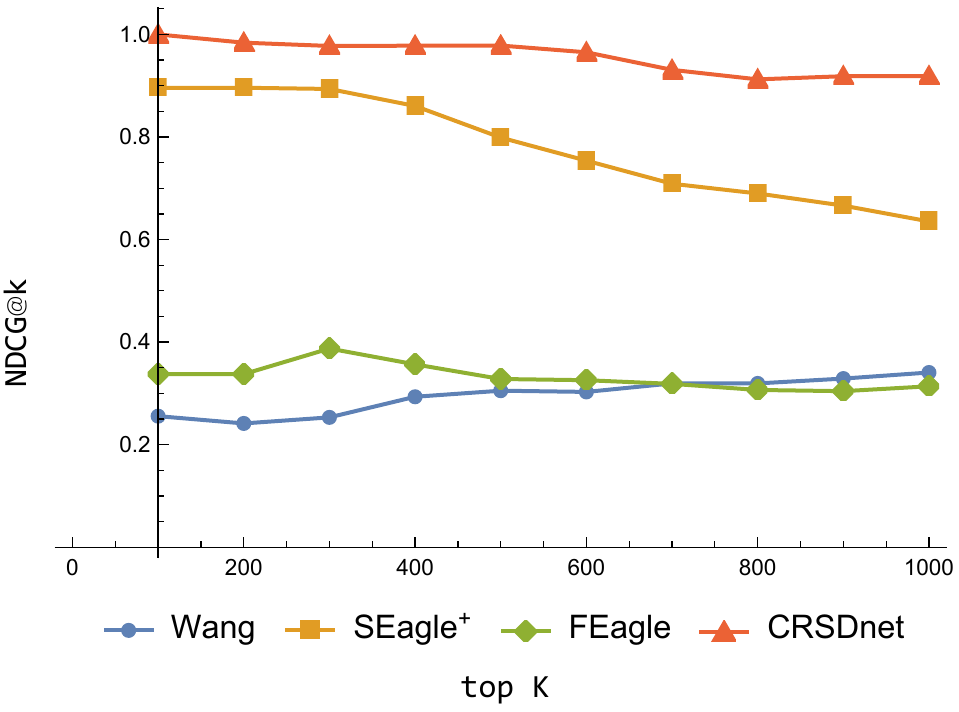}
  \caption{NDCG@k of compared methods on  YelpChi }
  \label{fig:NDCGYelpChi}
\end{figure}

\begin{figure}[htbp!]
  \centering
  \includegraphics[width=0.9\linewidth]{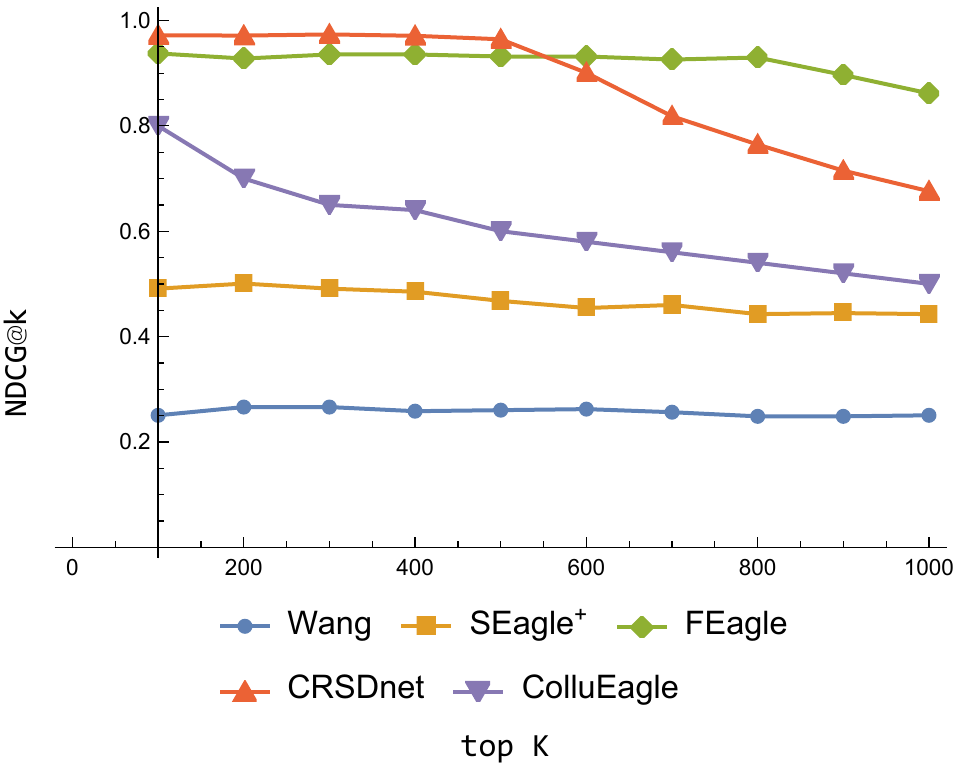}
  \caption{NDCG@k of compared methods on  YelpNYC }
  \label{fig:NDCGYelpNYC}
\end{figure}

\begin{figure}[htbp!]
  \centering
  \includegraphics[width=0.9\linewidth]{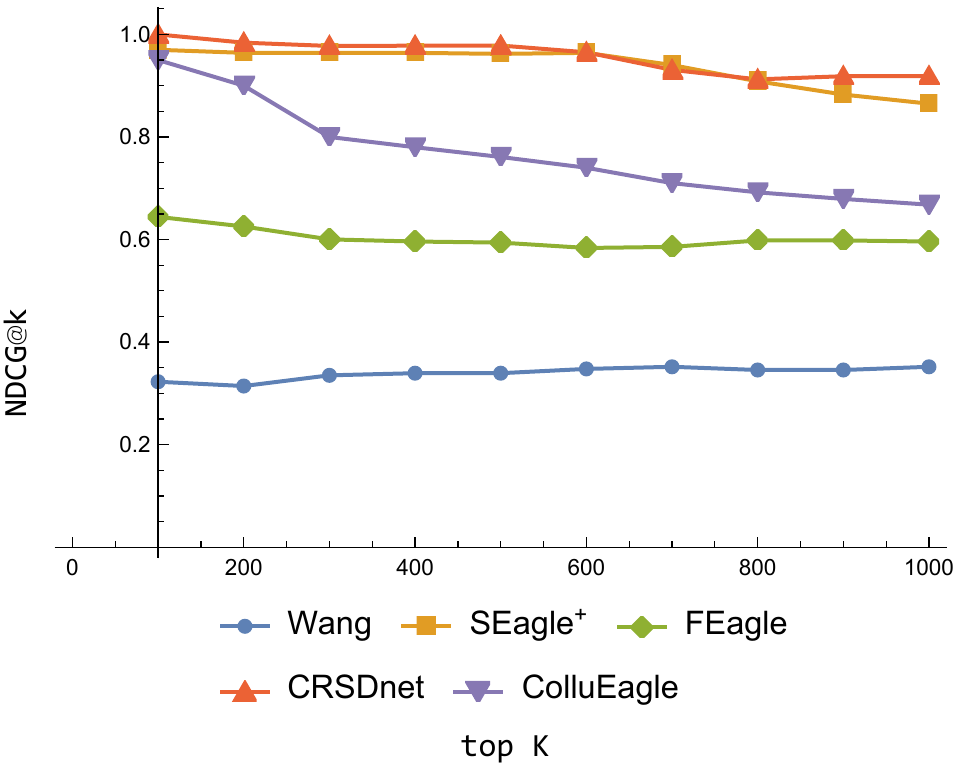}
  \caption{NDCG@k of compared methods on  YelpZip }
  \label{fig:NDCGYelpZip}
\end{figure}

Figures \ref{fig:NDCGYelpChi},\ref{fig:NDCGYelpNYC} and \ref{fig:NDCGYelpZip} plot the NDCG@k measure for $k$ between 0 and 1000. Here, all five competing algorithms are represented. Again, \mod~outperforms all algorithms for most values of $k$, in all three datasets.

Table \ref{tab:statDS} shows a comparison with two GNN-based approaches, \cite{9435380} and \cite{liu2020alleviating}. These approaches use much more data for training (80\%). As an additional baseline, we trained our Random Forest classifier this time on 30\% of the data, and predicted the reviewers' class. As evident from the table, more data is not a guarantee for better performance. One possible reason for the relative poor performance of NN-based methods is that much more data (in absolute value) is needed for successfully training the NN.


\section*{Conclusion}\label{section:conclusion}
In this work, we proposed a new holistic framework called
\mod~for detecting review spammers. Our method combines both machine learning and more classical algorithmic approaches (Belief Propagation) to better exploit the relational data (user–review– product) and metadata (behavioral and text data) to detect such users. Adding to previous work in this line of research, we come up with two new components: using machine learning to predict the edge and node potentials, and a new sampling rule in the active learning setting -- sample users from the largest clique. 

Our results suggest that the two components improve performance one on top of the other, and when combined, give the best result obtained so far for the Yelp dataset.

Another point that our work highlights is that while in many settings, NN-based methods give the best results, this is highly contingent upon having sufficient data for training. The spammer detection problem is exactly one of those problems where obtaining a lot of labeled data is expensive and non-trivial. Fake reviews are many times written by professionals, and it takes an experienced person to identify them. Hence platforms like Amazon Turk may not provide an easy solution to the shortage of data problem. In such cases, old-school algorithmic ideas become relevant again (Belief Propagation), and as we demonstrate in this paper, their performance may be boosted by incorporating ML in a suitable manner (computing potentials in our case) alongside domain expertise (sampling from the largest clique, following the insight about collusive spamming \cite{wang2020collueagle}).

One limitation of our work is the fact that we only tested on one platform, Yelp. Future work should run our pipeline on other datasets, once they become publicly available (as far as we know there is only one more dataset in English from Amazon which is publicly available). Also, we only considered the problem of user classification. It will be interesting to extend  our method  for the task of review classification.

\bibliography{ref.bib}
 
\appendix

\end{document}